\documentclass[11pt,a4paper]{article}
\usepackage[hyperref]{acl2019}
\usepackage{times}
\usepackage{latexsym}

\usepackage{url}

\usepackage{graphicx,color}
\usepackage{caption,todonotes}
\usepackage{amssymb}
\usepackage{amsmath}
\usepackage{multirow}
\usepackage{booktabs}

\aclfinalcopy 

\title{Dependency-Aware Named Entity Recognition with \\Relative and Global Attentions }

\author{Gustavo Aguilar \and Thamar Solorio \\
  Department of Computer Science \\
  University of Houston\\
  Houston, TX 77204-3010 \\
  \texttt{\{gaguilaralas, tsolorio\}@uh.edu}}

\date{}

\begin{document}
\maketitle
\begin{abstract}
Named entity recognition is one of the core tasks in NLP. Although many improvements have been made on this task during the last years, the state-of-the-art systems do not explicitly take into account the recursive nature of language. Instead of only treating the text as a plain sequence of words, we incorporate a linguistically-inspired way to recognize entities based on syntax and tree structures. Our model exploits syntactic relationships among words using a Tree-LSTM guided by dependency trees. Then, we enhance these features by applying relative and global attention mechanisms. On the one hand, the relative attention detects the most informative words in the sentence with respect to the word being evaluated. On the other hand, the global attention spots the most relevant words in the sequence. Lastly, we linearly project the weighted vectors into the tagging space so that a conditional random field classifier predicts the entity labels. 
Our findings show that the model detects words that disclose the entity types based on their syntactic roles in a sentence (e.g., verbs such as \textit{speak} and \textit{write} are attended when the entity type is \texttt{PERSON}, whereas \textit{meet} and \textit{travel} strongly relate to \texttt{LOCATION}).
We confirm our findings and establish a new state of the art on two datasets.
\end{abstract}

\section{Introduction}

\begin{figure*}
\centering
\includegraphics[width=\linewidth]{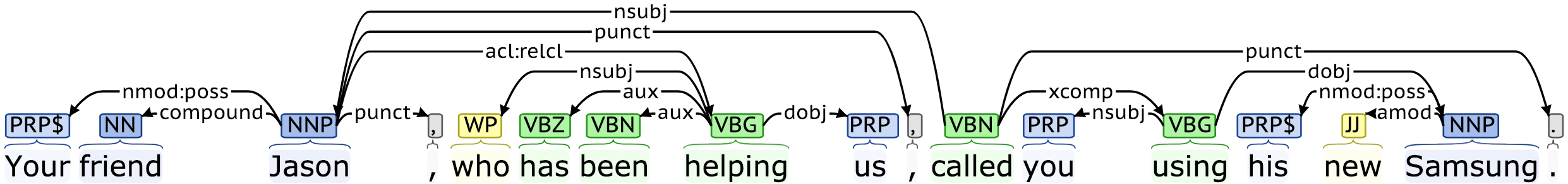}
\caption{ A dependency tree that shows how words relate among each other. It is worth noting that the connections \textit{called $\rightarrow$ Jason} and \textit{called $\rightarrow$ using $\rightarrow$ Samsung} capture how the entities \textit{Jason} and \textit{Samsung} interact. While the verb \textit{called} provides clues to the fact that \textit{Jason} is a person, the verb \textit{using} and the adjective \textit{new} suggest that \textit{Samsung} is a product and not a company.}
\label{fig:dependency_tree}
\end{figure*}

In the study of the named entity recognition (NER) task, neural sequence labeling models have been vastly explored by the NLP community. Small variations aside, the majority of these models use a combination of a bidirectional LSTM and conditional random fields (CRF) to reach state of the art performance \citep{DBLP:journals/corr/LampleBSKD16, DBLP:journals/corr/MaH16, DBLP:journals/corr/PengD16, yang2018design, aguilar-EtAl:2018:N18-1}. Recently, transfer learning from pre-trained language models has played an important role on improving the performance even further \citep{DBLP:journals/corr/abs-1709-04109, peters-EtAl:2018:N18-1, C18-1139, devlin2018bert, DBLP:journals/corr/abs-1801-06146}. However, these NER models mainly consider the text as a plain sequence of words without explicitly taking into account the recursive nature of language. Consider the phrase in Figure \ref{fig:dependency_tree}, ``\textit{Your friend Jason, who has been helping us, called you using his new Samsung.}'' 
The main sentence is composed of a verb phrase whose head is \textit{called}, which in turn includes another verb phrase headed by the word \textit{using}.
In addition to this, such recursive properties can produce more involved and longer sentences. This behavior can potentially obscure relationships among words when the text is treated as a linear chain of tokens. For instance, in the same example, the words \textit{Jason}, \textit{called}, \textit{using}, and \textit{Samsung} easily describe the way in which the entities \textit{Jason} and \textit{Samsung} interact\footnote{Some verbs provide sufficient clues to determine the entity type of the subject that performs the action. }. 
Nevertheless, existing models struggle identifying such relationships in long sentences, resulting in a drop in performance.

We propose a new approach for NER where the goal is to enhance the syntactic relationships among words and combine such aspects with semantic representations commonly used in this task. Our model extracts features from text using a Tree-LSTM 
\cite{tai-socher-manning:2015:ACL-IJCNLP} guided by dependency tree structures. As shown in Figure \ref{fig:dependency_tree}, the dependency trees connect the words based on the role they play in the sentence and the way they interact with each other. The output features are weighted with relative and global attention mechanisms. While the relative attention focuses on the most relevant words with respect to the word being evaluated, the global attention spots the most important words over the sentence as a whole. After weighting the hidden vectors, we linearly project them into the tagging space. We predict the entity labels using a conditional random field classifier.
Our findings show that the model detects words that disclose the entity types based on their syntactic roles in a sentence (e.g., verbs such as \textit{speak} and \textit{write} are attended when the entity type is \texttt{PERSON}, whereas \textit{meet} and \textit{travel} strongly relate to \texttt{LOCATION}).

The contribution of this work can be summarized as follows: we (a) apply global and relative attentions to detect informative word relationships, (b) provide an end-to-end architecture that involves language model, Tree-LSTM, attention mechanisms, and CRF, (c) derive insights from entity and word correlations learned by the model, (d) release the entire project that allows replicability of all the described experiments and analysis, and (e) establish a new state of the art on the OntoNotes 5.0 Broadcast News section, and SemEval 2010 Task 1 datasets.

\section{Related Work}

Recursive neural networks started to be considered in computer vision to target the compositionality problem. \citet{Socher:2011:PNS:3104482.3104499} combined fragments of images associated to constituencies in binary parsing trees to extract composed features using recursive neural networks. However, the main problem was that their model did not retain the information along the tree but it only captured the immediate comparison between adjacent nodes. 
Matrix-vector models aimed at keeping long term information with a reasonable performance \cite{Socher:2012:SCT:2390948.2391084}, but \citet{tai-socher-manning:2015:ACL-IJCNLP} proposed a more feasible way to overcome the compositionality problem. They introduced the multi-purpose Tree-LSTM architecture that generalizes the standard LSTM \cite{doi:10.1162/neco.1997.9.8.1735}. The Tree-LSTM model is capable to handle N-ary trees using dynamic batching.
They also improved the state of the art in sentiment analysis at that point in time, which shows the potential of modeling linguistic problems recursively. 
We follow the implementation of Child-Sum Tree-LSTM to extract syntactic features guided by dependency trees.

Dependency parses have been hardly employed on NER. Recently, \citet{AAAI1714741} explored this line of research with slightly different focus. They concentrate on efficient models that show less running time by using a semi-Markov CRF to exploits dependency features. We compare our model with their work since their approach also extracts features using Tree-LSTM on dependency trees. 
Additionally, \citet{jointRLandDLforNERandREL} proposed a deep reinforcement learning framework where the goal is recognize entities and extract their relations simultaneously. They identify a candidate pair of entities using LSTM and an attention mechanism, then they employ a Tree-LSTM to extract the entity relation using the path between the pairs in a dependency tree. They use this information in a Q-learning algorithm to optimize the policy of their system. Unlike them, we use the dependency trees to extract syntactic features that describe the interaction between entities and the rest of the words, regardless if they are other entities. This allows us to find words that disclose entity type information according to the sentence syntactic structure (e.g., verbs that only \texttt{PERSON} entities perform).

We enhance our syntactic features using two different attention mechanisms. Attention was introduced by \citet{DBLP:journals/corr/BahdanauCB14} in the task of machine translation. Since then, it has been broadly used in many other applications such as semantic slot filling \citep{D17-1004} and sentiment analysis. We employ a similar attention component, which we call global attention (see Section \ref{sec:global_attention}).
More recently, \citet{DBLP:journals/corr/VaswaniSPUJGKP17} proposed the transformer architecture, which contains a multi-head self-attention module. Our relative attention mechanism is very similar to this module, but the main difference is that, for a word $w_i$ in a sentence, we draw the probability distribution over the words $w_j$ where $i \neq j$ (see Section \ref{sec:relative_attention}). Additionally, the transformer architecture has been used in the language model BERT \cite{devlin2018bert}, which is benchmarked on NER. 

NER has evolved rapidly in the last years. While the standard architectures involve a combination of BLSTMs with a CRF classifier \cite{DBLP:journals/corr/LampleBSKD16, LimsopathamAndCollier:16, Chieu:2003:NER, peng-dredze:2016:P16-2, DBLP:journals/corr/ChiuN15, aguilar-EtAl:2018:N18-1}, recent advances have improved those approaches with transfer learning from pretrained language models such as ELMo \cite{peters-EtAl:2018:N18-1} and BERT \cite{devlin2018bert}. These language models show a similar performance on the CoNLL 2003 benchmark, but the current state of the art on this dataset is the contextual string embeddings from language models proposed by \citet{C18-1139}. We take advantage of such improvements by using ELMo to generate our contextualized word representations. 
\section{Methodology}
This section describes the feature representation, the model architecture, and the training details.

\subsection{Feature representation} \label{sec:feat_representation}

We represent the input data using words, part-of-speech tags, and dependency parses. For words, we employ deep contextualized representations using the language model ELMo 
\citep{peters-EtAl:2018:N18-1}. ELMo provides vector representations that are entirely built out of characters. This allows us to overcome the problem of out-of-vocabulary words by always having a vector based on morphological clues for any given token. For POS tags and dependency relations, we use trainable embedding matrices that are optimized from scratch. POS tags have proven useful in previous research \cite{DBLP:journals/corr/ChiuN15, aguilar-EtAl:2018:N18-1}, and dependency relations help the model to infer the interaction between nodes in the trees. Once we have a vector representation for every input feature, we concatenate them to form a single vector for every token in the sentence.

\subsection{Model architecture}

We describe the components of our model individually, and then we discuss the overall architecture.

\subsubsection{Tree-LSTM}

The Tree-LSTM component, introduced as Child-Sum Tree-LSTM by \citet{tai-socher-manning:2015:ACL-IJCNLP}, is a generalization of the standard LSTM cell \cite{doi:10.1162/neco.1997.9.8.1735} that can handle multiple inputs at every time step. In fact, both cells are equivalent when the input tree is comprised of a single child at every level\footnote{The Child-Sum Tree-LSTM runs in a bottom up fashion, which makes it equivalent to a reversed LSTM when the root is the first word in a plain sequence of tokens.}. The equations are as follows:
\begin{equation*} \label{eq:treelstm_cell}
\begin{split}
\tilde{h}_j =& ~\sum_{k \in C(j)}h_k  \\
i_j         =& ~\sigma(\mathrm{W}^{(i)} x_j + \mathrm{U}^{(i)} \tilde{h}_j + b^{(i)})  \\
f_{jk}      =& ~\sigma(\mathrm{W}^{(f)} x_j + \mathrm{U}^{(f)} h_k + b^{(f)})  \\
o_j         =& ~\sigma(\mathrm{W}^{(o)} x_j + \mathrm{U}^{(o)} \tilde{h}_j + b^{(o)})  \\
u_j         =& ~\mathrm{tanh}(\mathrm{W}^{(u)} x_j + \mathrm{U}^{(u)} \tilde{h}_j + b^{(u)})  \\
c_j         =& ~i_j \odot u_j + \sum_{k \in C(j)} f_{jk} \odot c_k  \\
h_j         =& ~o_j \odot \mathrm{tanh}(c_j) \\
\end{split}
\end{equation*}
\noindent where $k \in C(j)$, and $C(j)$ determines the children of the node $j$. The differences are that, in a given time step (i.e., node), the forget cell $f_{jk}$ is calculated for every hidden input state $h_k$, and the cell state $c_j$ is the sum of the products between $f_{jk}$ and $c_{k}$ for every child $k$ of node $j$. In our case, the dependency trees consist of words at every node. That is, the input $x_j$ is an embedding vector representing a word $w_j$. Its children are the words $w_k$ which produce the $h_k$ hidden states. Since children are used as input to the next node, the Tree-LSTM model runs in a bottom-up fashion. In the case of multi-rooted phrases, we simply average the root vectors to come up with a single last hidden state $h_{root}$.

\subsubsection{Relative attention} \label{sec:relative_attention}

The goal of the relative attention is to produce a Cartesian product of probabilities based on the words in the sentence, where the main diagonal is not taken into account to draw the probability distribution. This is slightly different from the self-attention mechanism proposed by \citet{DBLP:journals/corr/VaswaniSPUJGKP17} in that we draw the probability distribution over the words $w_j$ of the sentence for every word $w_i$, where $i \neq j$, as opposed to distributing the probabilities over all the words regardless. We use the scaled-dot product function to produce the attention matrix. That is, we scale the weights by the inverse square root of their embedding dimension, mask out the main diagonal of such scores, and normalize the result using a softmax function:
\begin{equation} \label{eq:relative_attention_matrix}
A =~ \mathrm{softmax}(d^{-0.5}_a QK^T)
\end{equation}

\noindent where $A \in \mathbb{R}^{N \times N}$ is a squared matrix that contains the attention weights for $N$ words in a sentence, $i$ and $j$ denote the row and column indexes, such that $\sum_{i}^N\sum_{j \neq i}^N A_{ij} = N$. 
$Q$ and $K$ are linear transformations of the input using the query and key matrices $\mathrm{W}_Q \in \mathbb{R}^{d_a \times d_a}$ and $\mathrm{W}_K \in \mathbb{R}^{d_a \times d_a}$ where $d_a$ is the dimension of the input and output matrices. The weighted values are calculated as follows:
\begin{equation}
M =~ A V + V
\end{equation}

Similar to $Q$ and $K$, $V$ is a linear projection of the input using the value matrix $\mathrm{W}_V \in \mathbb{R}^{d_a \times d_a}$. Note that the matrix multiplication between $A$ and $V$ discards the words $w_{ii}$ because $A$ contains zeros in its main diagonal. Hence, we include this information by adding the matrix $V$ to such product\footnote{This is equivalent to $M = (A+I)V$ where $I \in \mathbb{R}^{N \times N}$ is the identity matrix.}. 

\subsubsection{Global attention} \label{sec:global_attention}

In the case of the global attention, we use a fairly standard mechanism introduced by \citet{DBLP:journals/corr/BahdanauCB14}. The idea is to concentrate mass probability over the words that capture the most relevant information along the sentence. Our attention mechanism uses the following equations:
\begin{equation} \label{eq:attention-context}
	\begin{split}
		u_i =&~ v^\intercal ~\mathrm{tanh}(\mathrm{W}_h h_i + b_h + \mathrm{W}_q q + b_q) \\
		a_i =&~ \frac{\mathrm{exp}(u_i)}{\sum^{N}_{j=1}\mathrm{exp}(u_j)}, ~~~\text{s.t.}~\sum_{i=1}^N a_i = 1 \\
		z_i =&~ a_i h_i
	\end{split}
\end{equation}
where 
$\mathrm{W}_h \in \mathbb{R}^{d_a \times d_h}$, 
$\mathrm{W}_q \in \mathbb{R}^{d_a \times d_q}$, 
$b_h \in \mathbb{R}^{d_a}$, and 
$b_q \in \mathbb{R}^{d_a}$
are learnable parameters of the model. $\mathrm{W}_h$ and $\mathrm{W}_q$ are used to linearly project the hidden word vectors $h_i$ and the query vector $q$ into the attention space. The vector $v \in \mathbb{R}^{d_a}$ is the attention vector to be learned. $d_h$, $d_q$, and $d_a$ are the dimensions of the hidden word vectors, the query vector, and attention layer, respectively. Note that the query vector $q$ is a context vector that summarizes the entire sentence. In the case of using a Tree-LSTM, $q$ is the root hidden vector $h_{root}$, whereas in a LSTM $q$ is simply the last hidden state $h_n$. Finally, we multiply the scalars $a_i$ and their corresponding hidden vectors $h_i$ to obtain our weighted sequence $z$. 
 
\subsubsection{Residual connections}

We incorporate residual connections \cite{DBLP:journals/corr/HeZRS15} at every component of our module, followed by layer normalization as in \citet{ba2016layer}. The output of a given $\mathrm{Sublayer}$ is described as follows:
\begin{equation}
z = \mathrm{LayerNorm}(x + \mathrm{Sublayer}(x))
\end{equation}

\noindent where $\mathrm{LayerNorm}$ is an affine function that contains trainable parameters. Additionally, $\mathrm{Sublayer}$ can be any component of our model, such as a Tree-LSTM, relative attention, or global attention. We keep the same dimensions for inputs and outputs to simplify adding the vectors of any given module. This module only normalizes the output tensor in the last dimension.

\subsubsection{Conditional random field}

We use a conditional random field (CRF) classifier at the top of our model to perform the sequential inference. The CRF takes vectors in the tagging space as input and produces the best sequence of labels using the Viterbi algorithm. CRF is well-known and widely used for sequence labeling because it learns the rules of transitioning from one label to another based on the feature vectors of the sequence as a whole instead of individually. 

Consider the observation sequence of vectors $\mathbf{x} = [x_1, \dots, x_n]$ and its corresponding target labels $\mathbf{y} = [y_1, \dots, y_n]$. CRF computes the conditional probability of the target sequence $\mathbf{y}$ given the inputs $\mathbf{x}$ by globally normalizing the target score: 
\begin{equation} \label{eq:crf_cond_prob}
\begin{split}
    p_\theta(\mathbf{y}|\mathbf{x}) =&~ \frac{1}{\mathbf{Z}_\theta(\mathbf{x})}\prod_{t=1}^{N} \psi(y_{t-1}, y_t, \mathbf{x};\theta)
\end{split}
\end{equation}
\noindent where $\mathbf{Z}_{\theta}(\mathbf{x})$ is a normalization term that adds up the products of $\psi(\cdot)$ for all the possible $\mathbf{y}$ sequences.
$\psi(\cdot)$ is the potential parametric function that sums the transition and emission features. We use the logistic expression of Equation \ref{eq:crf_cond_prob} during training to optimize our model (see Section \ref{sec:training}).
\subsubsection{Overall architecture}

\begin{figure}
\centering
\includegraphics[width=0.97\linewidth]{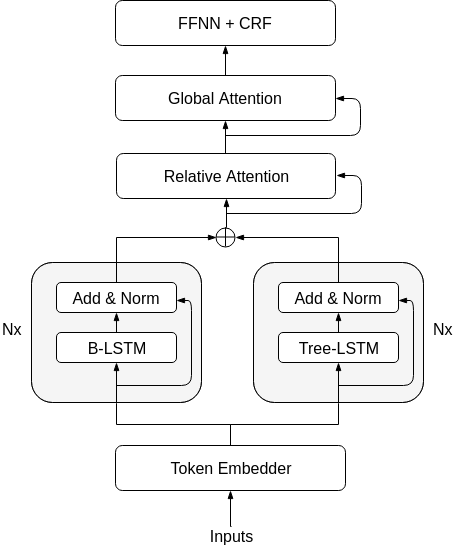}
\caption{ The overall model architecture. }
\label{fig:overall_model}
\end{figure}

In this section we describe the overall architecture of our model using the previous components. As shown in Figure \ref{fig:overall_model}, we first embed the input sentence into a vector space using the token embedder module. This module is in charge of concatenating the word, POS tag, and dependency relationship vectors into a single representation for every token (see Section \ref{sec:feat_representation}). We feed the embedded output into the semantic (on the left) and syntactic (on the right) feature extractors, which are a stacked layers of bidirectional LSTMs and Tree-LSTMs, respectively. Then, we concatenate the outputs of these components and feed them into the relative and global attention modules. The weighted vectors generated by the attention mechanisms are linearly projected into the tagging space and fed into a conditional random field classifier. The arrows that skip layers denote the residual connections.

\subsection{Training} \label{sec:training}

We optimize our model by minimizing the negative log-likelihood loss produced by the forward algorithm in the CRF module. In addition to this loss, we include an $\ell_2$ regularization term that targets the parameters of the semantic and syntactic feature extractors and the attention mechanisms. The idea is to reduce over-fitting when the model selects the features from the syntactic and semantic blocks, and to force the attention mechanisms to avoid bias towards specific aspects of the sentences. Given the target sequence $\mathbf{y}$ and the predicted labels $\mathbf{\hat{y}}$, we define the objective function\footnote{This function expresses the loss for a single input sequence, but in practice we average the sum of all the token-level losses across the batch sequences.} of our model as follows:
\begin{align}
	\mathcal{L} =&~ -\frac{1}{N}\sum_i^N y_{i} log(\hat{y}_{i}) + \lambda \sum_k{w_k^2}
\end{align}

\noindent where $N$ is the length of the sentence and $w_k$ denotes the parameters of the feature extractors and the attention mechanisms. $\lambda$ is the penalty that indicates how much of this regularization term will be added to the overall loss, and $log(\hat{y_i})$ is determined by the logistic expression from Equation \ref{eq:crf_cond_prob}.


We reduce the loss of our model using Stochastic Gradient Descent (SGD) with momentum \cite{pmlr-v28-sutskever13}. We train our models for 150 epochs and change the learning rate every epoch using cosine annealing and warm restarts \cite{DBLP:journals/corr/LoshchilovH16a}. Besides adding $\ell_2$ regularization, we also prevent over-fitting by applying input variational dropout \cite{Gal2015DropoutB} to every component in our model. Appendix \ref{app:sec:hyperparameters} includes all the details regarding hyperparameter tuning.

\begin{table}[t!]
\centering
\small
\begin{tabular}{lllll} 
\toprule
\textbf{Corpus}     & \textbf{Train}    & \textbf{Dev}   & \textbf{Test} & \textbf{NEs}     \\ \midrule 
CoNLL 2003          &  14,041           &  3,250                & 3,453         & 4         \\
SemEval 2010        &  3,648            &  741                  & 1,141         & 22        \\ 
CoNLL 2012          &  115,310          &  15,680               & 12,217        & 18        \\
OntoNotes 5.0       &  9,723            &  1,172                & 1,252         & 18        \\
\bottomrule
\end{tabular}
\caption{The data splits on each corpus along with the number of named entities.}
\label{tab:dataset_splits}
\end{table}

\section{Experimental results}

\begin{table}[t!]
\centering
\small
\begin{tabular}{lllll} 
\toprule
\textbf{Model Approach} & \textbf{CoNLL} & \textbf{SemEval} & \textbf{ON 5.0} \\ 
\textbf{}& \textbf{2003}  & \textbf{2010}    & \textbf{(BN)} \\ 
\midrule
\textit{Baselines}                              &   &   &  \\
~\textsc{Tlstm\textsubscript{Word}}             & 86.72	    & 79.53     & 82.20     \\
~\textsc{Tlstm\textsubscript{Word+POS}}         & 88.91	    & 84.07     & 88.41     \\
~\textsc{Tlstm\textsubscript{Word+POS+Dep}}     & 89.20	    & \bf 84.89 & \bf 88.67 \\
~\textsc{Blstm\textsubscript{Word}}             & 89.06	    & 80.81     & 83.30     \\
~\textsc{Blstm\textsubscript{Word+POS}}         & \bf 91.22	& 80.31     & 84.58     \\
~\textsc{Blstm\textsubscript{Word+POS+Dep}}     & 90.98	    & 81.23     & 83.24     \\ \midrule 
\textit{Stacked Layers}                         &   &   &  \\
~\textsc{Tlstm}                                 & 91.22	    & 85.01	    & 88.08     \\ 
~\textsc{Tlstm}$\dag$                          & 91.24	    & \bf 85.75	& 88.71     \\
~\textsc{Blstm}                                 & 91.68	    & 84.32	    & 88.48     \\ 
~\textsc{Blstm}$\dag$                          & \bf 91.71	& 85.21	    & \bf 88.77 \\ \midrule  
\small{\textit{Attention}}                      &   &   &  \\
~\textsc{Tlstm + RA}$\dag$                     & 91.13     & 84.94	    & 88.11     \\
~\textsc{Tlstm + GA}$\dag$                     & 90.62     & 85.09	    & \bf 89.02 \\ 
~\textsc{Tlstm + RA + GA}$\dag$                & 90.60     & \bf 86.24	& 88.79     \\ 
~\textsc{Blstm + RA}$\dag$                     & \bf 91.65 & 84.19	    & 88.51     \\
~\textsc{Blstm + GA}$\dag$                     & 91.39     & 84.87	    & 88.39     \\ 
~\textsc{Blstm + RA + GA}$\dag$                & 91.63     & 85.12	    & 88.66     \\ \midrule
~\textsc{Tlstm + Blstm}                         & 91.17    & \bf 86.49 & \bf 89.22 \\
~\textsc{+ RA + GA}$\dag$                      &           &           &           \\
\bottomrule
\end{tabular}
\caption{The averaged F1 scores of two runs on the validation sets.
\textsc{Blstm} and \textsc{Tlstm} are \textbf{b}idirectional and \textbf{t}ree LSTMs. The baselines specify the input features; the rest of the experiments use all the features. $\dag$ denotes residual connections on every component. \textsc{RA} and \textsc{GA} refer to relative and global attentions. All the experiments use ELMo embeddings and CRF.}
\label{tab:devset_results}
\end{table}

\begin{table}[t!]
\centering
\small
\renewcommand{\arraystretch}{1.1}
\setlength{\tabcolsep}{2.5pt}
\begin{tabular}{lllll} 
\toprule
\textbf{Dataset}    & \textbf{Last F1 SOTA}      & \textbf{Our F1}       \\ \midrule 
\multirow{3}{*}{CoNLL-03}
    & 93.18 \cite{C18-1139}      & 91.90$\pm$0.08  \\
    & 92.8 \cite{devlin2018bert}      &   \\
    & 92.22 \cite{peters-EtAl:2018:N18-1}      &   \\ 
\midrule

SemEval-10          & 75.50 \cite{AAAI1714741}   & \bf 86.72$\pm$0.11   \\ 
\midrule
\multirow{2}{*}{CoNLL-12}            
    & 89.30 \cite{C18-1139}      & 89.07$\pm$0.15      \\
    & 86.28 \cite{DBLP:journals/corr/ChiuN15} & \\ 
\midrule
ON 5.0 (BN)         & 80.50 \cite{AAAI1714741}   & \bf 89.22$\pm$0.28  \\
\bottomrule
\end{tabular}
\caption{Comparison of our best F1 results and the state of the art reported on the test set of each corpus. 
Our model exceeds previous scores on SemEval 2010 and the OntoNotes 5.0 (BN), and it is competitive with the scores on CoNLL 2012.}
\label{tab:testset_results}
\end{table}

We run experiments on CoNLL 2003 \cite{TjongKimSang-DeMeulder:03}, SemEval 2010 Task 1 \cite{recasens2010semeval}, CoNLL 2012 \cite{pradhan2013towards}, and the Broadcast News section of OntoNotes 5.0 \cite{ontonotes5}. Table \ref{tab:dataset_splits} shows a summary of the data splits for each corpus along with the number of entities (more details in Appendix \ref{app:sec:datasets}). 
The scores on the test set for each dataset are reported in Table \ref{tab:testset_results}. 

\subsection{Ablation analysis}

Table \ref{tab:devset_results} shows a systematic way to incorporate the components of our model. Based on those results, we can infer the impact of adding a component to the final architecture.
For the input features, we observe that adding POS tags improves significantly the performance compared to only using the words in either \textsc{Blstm} or \textsc{Tlstm}. This is consistent with previous research (see Section \ref{sec:feat_representation}). An additional small improvement can be achieved by incorporating the dependency relation labels. It is worth noting that, while the \textsc{Blstm\textsubscript{Word}} or \textsc{Tlstm\textsubscript{Word}} are similar in performance, the latter significantly overpasses the former when POS tags are added. Intuitively, the POS tags refine the syntactic patterns extracted from the tree structures and clarify the roles that the words play in the sentence.

For the model components, we experiment with stacked layers on the \textsc{Blstm} and \textsc{Tlstm} models (see Table \ref{tab:devset_results}). 
While there is no significant improvement over the single-layer models, we see gains when we add residual connections between each layer. The residual connections allow the model to decide whether it needs to go deeper or just skip the stacked layers at its convenience. Additionally, we show the performance of the models when the relative and global attention mechanisms are added separately. However, they only improve the model significantly when both are combined.

\subsection{Attention analysis}

\begin{figure}[t!]
\centering
\includegraphics[width=\linewidth]{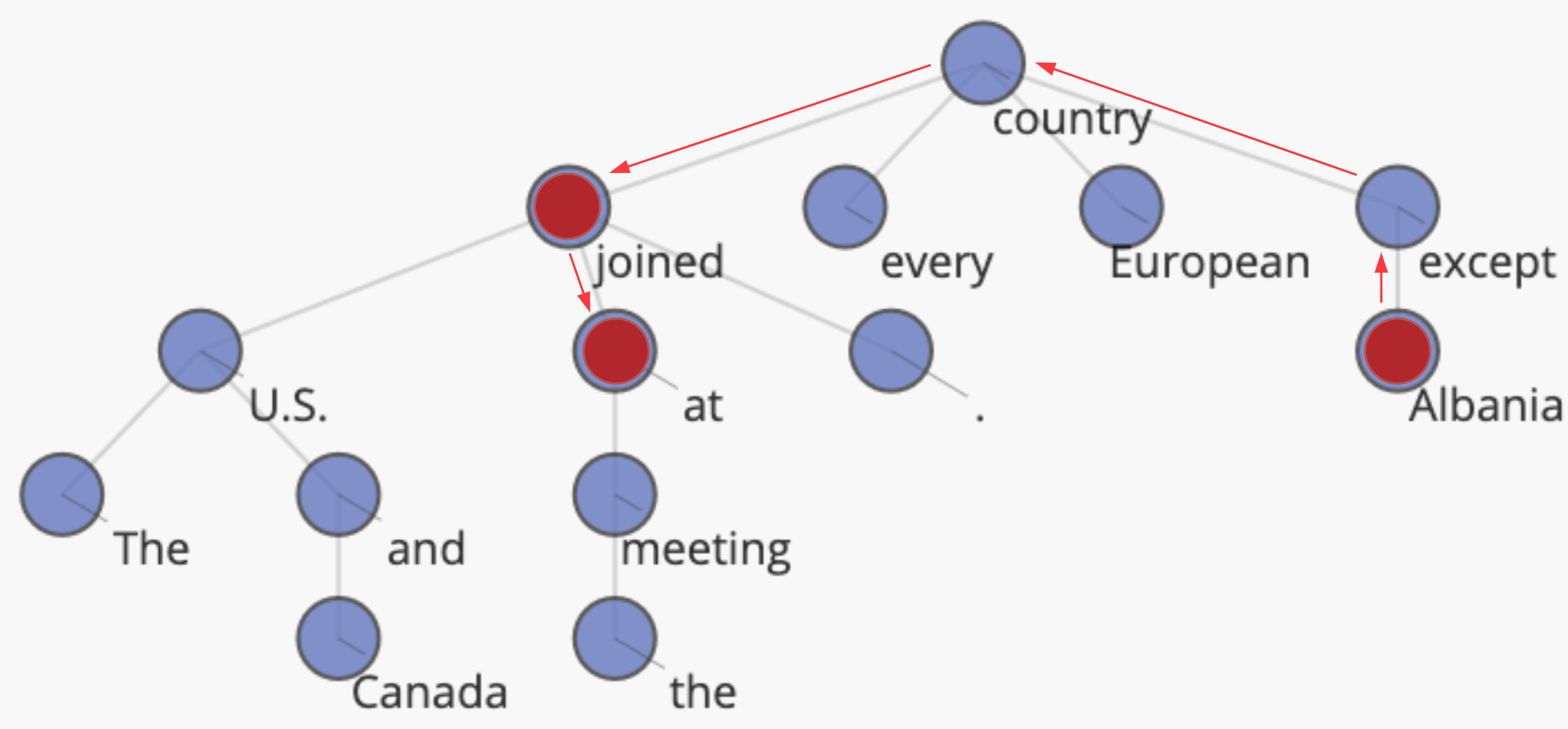}
\caption{ The dependency tree of the sample sentence. We highlight the path from the word \textit{Albania} to its most attended words \textit{joined} and \textit{at}.}
\label{fig:rel_attn_tree_semeval10}
\end{figure}

\begin{figure}[t!]
\centering
\includegraphics[width=\linewidth]{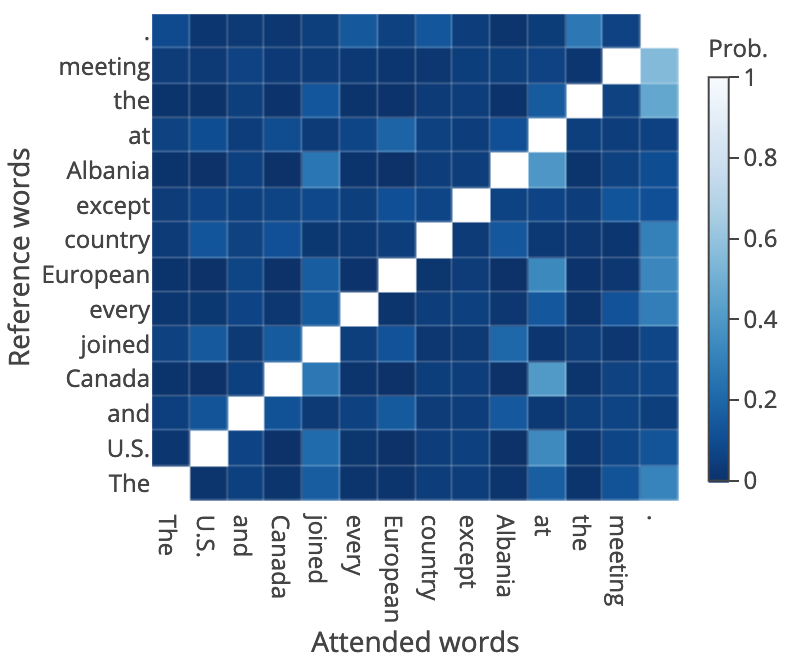}
\caption{ The relative attention matrix $A$, as described in Eq. \ref{eq:relative_attention_matrix}, holds a probability distribution on every row. The main diagonal is excluded from all the probability distributions.}
\label{fig:rel_attn_semeval10}
\end{figure}

\begin{figure*}[t!]
\centering
\includegraphics[width=\linewidth]{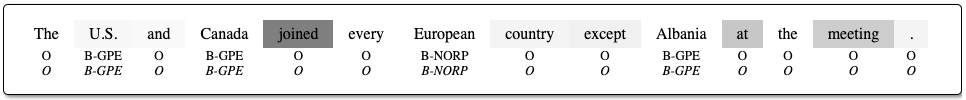}
\caption{ The visualization of global attention. The more highlighted the token is, the more probability mass it has. The labels and predictions (italics) appear below each word.}
\label{fig:glob_attn_semeval10}
\end{figure*}

\noindent \textbf{Relative attention}. Figure \ref{fig:rel_attn_semeval10} shows the relative attention matrix for the dependency tree in Figure \ref{fig:rel_attn_tree_semeval10}. Each row of the matrix draws a probability distribution over the rest of the words (columns). By inspecting the matrix, it is easy to note that the words \textit{at} and \textit{joined} in the x-axis are the most relevant for \textit{Albania} in the y-axis, whereas \textit{U.S.}, \textit{Canada}, \textit{European}, and \textit{Albania} (x-axis) are the most important for \textit{joined} (y-axis). This supports the idea that prioritizing words based on their relationships can provide different perspectives of the sentence at the word level. Additionally, when we inspect the models \textsc{Blstm} and \textsc{Tlstm} separately, we see different patterns captured in their attention matrices; for \textsc{Blstm} the relations are prioritized semantically 
(e.g., verbs are the most important parts and prepositions are hardly highlighted), whereas for \textsc{Tlstm} it is more syntactically (e.g., prepositions, verbs, punctuations are all relevant). Merging both techniques improves the results because they are complementary.

\noindent \textbf{Global attention}. Figure \ref{fig:glob_attn_semeval10} shows the attended words once the relative attention matrix has extracted the prioritized relations. The figure shows a combination of both syntactic and semantic patterns. That is, while words like \textit{joined} and \textit{meeting} would be semantically expected, the model also focus on \textit{except}, \textit{at}, and the punctuation. Importantly, our model does not act as an entity spotter at this level\footnote{e.g., \textsc{Blstm\textsubscript{Word}} with global attention tends to highlight entities suggesting memorization rather than generalization}. Instead, it relies on the 
syntactic structure 
of the sentence. Additionally, note that the word \textit{meeting} was not too relevant in the relative attention matrix, but at the global perspective, it becomes important. This suggests that the attention mechanisms are complementary.

\begin{figure}
\centering
\includegraphics[width=\linewidth]{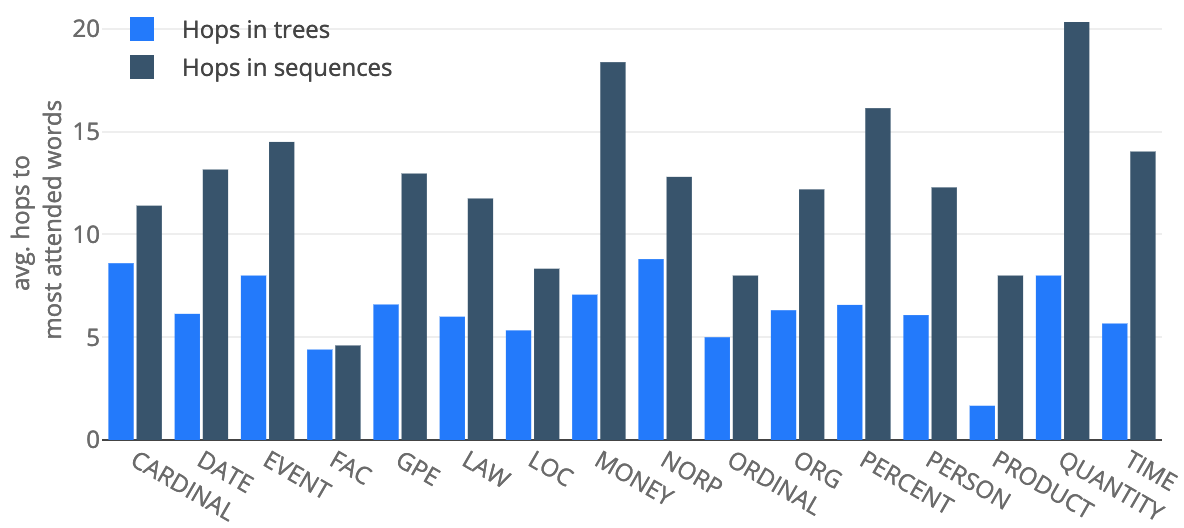}
\caption{ Average of hops to the most attended words in the validation set with respect to the entity tokens. For sequences, the hops are the number of tokens in between the entity and the most attended word. For trees, the hops are the number of nodes in the path that connects the entity and the most attended word.}
\label{fig:hops_comparison_semeval10}
\end{figure}


\begin{table*}[t!]
\centering
\small
\setlength{\tabcolsep}{8pt}
\begin{tabular}{lllp{10cm}}
\toprule
\textbf{Entity}     & \textbf{POS}  & \textbf{Coverage}  & \textbf{Most Attended Words} \\ \midrule 
\multirow{8}{*}{\texttt{PERSON}}
    & \multirow{4}{*}{\texttt{NN}}       
        & \multirow{4}{*}{31.53\%}   
        & \textit{mr., correspondent, events, assistant, suit, case, ms., poll, a.m., consultant, president, express, officer, ignorance, cbs, wife, member, years, diet, menswear, life, relations, sunday, instance, season, network, school, woman, professor} \\ \cmidrule{2-4}
    & \multirow{4}{*}{\texttt{VB}}       
        & \multirow{4}{*}{15.88\%}   
        & \textit{say, speak, come, go, write, begin, produce, need, tie, star, schedule, re-create, report, take, die, call, feature, love, continue, agree, know, pinch, consider, co-anchored, waive, brush, happen, issue, thank, welcome, slat, age, loot, resign, wrap, fall, chew, entomb, upset, allay, admire, place, strengthen} \\ \midrule
\multirow{8}{*}{\texttt{LOCATION}}
    & \multirow{2}{*}{\texttt{VB}}       
        & \multirow{2}{*}{28.42\%}   
        & \textit{show, mellow, hold, direct, entitle, possess, continue, allow, travel, reach, begin, report, play, meet, carry, cause, miss, reduce, happen, mean, threaten, decide} \\ \cmidrule{2-4}
    & \multirow{4}{*}{\texttt{NN}}       
        & \multirow{4}{*}{23.50\%}   
        & \textit{approaches, pollution, correspondent, temperatures, earthquake, u.s., meters, toepfer, run, re-enactment, behavior, mechanisms, breach, secrets, francisco, suspect, verdict, role, clue, analysis, thompson, space, story, zealand, violence, freedom, europe, front, glacier, conference, way, russia, death, cause, lake} \\
\bottomrule
\end{tabular}
\caption{The most attended words for nouns that are labeled as \texttt{PERSON} or \texttt{LOCATION}. The table shows the attended POS tag per entity, its coverage, and the corresponding list of words. 
}
\label{tab:ne_correlations}
\end{table*}

\noindent \textbf{Attended word distances}. We also calculate the distance between the entities and its most attended words. That is, we take the attention matrix $A$, look for the rows whose tokens are labeled as entities, extract the most attended words, and calculate the number of hops between the entities and the extracted words for both the sentence sequence and the dependency trees. Figure \ref{fig:hops_comparison_semeval10} clearly shows that the most attended words by our model are closer when their layout is a tree instead of a sequence. This partially explains the better performance of the \textsc{Tlstm\textsubscript{Word+PoS}} over the \textsc{Blstm\textsubscript{Word+PoS}}, which also aligns with our claim that the long distance relations are potentially lost along the sentence when the words are treated as a plain sequence of tokens.

\noindent \textbf{Entity correlations}. We investigate whether the attended words with respect to the entity tokens follow any particular pattern. We conduct the study by controlling over the entity types and their corresponding POS tags. Specifically, we only consider the entity tokens whose POS tags are nouns (e.g., \texttt{NN}, \texttt{NNS}, \texttt{NNP}, or \texttt{NNPS}). Then, we extract the top three most attended words along with their POS tags from the SemEval 2010 validation set. In Table \ref{tab:ne_correlations}, we show the coverage of the most attended POS tags and their corresponding words for \texttt{PERSON} and \texttt{LOCATION}. 
For the type \texttt{PERSON}, the nouns \textit{president}, \textit{assistant}, and \textit{officer} are roles that only people perform, which easily discriminate the type \texttt{PERSON} from any other entity type. In the case of its attended verbs, \textit{speak}, \textit{love}, and \textit{die} are actions also performed by people (i.e., the entity is the subject of the sentence). For the type \texttt{LOCATION}, the verbs \textit{entitle}, \textit{travel}, and \textit{meet} appear in cases where the location is the object of the sentence, and as such, these verbs disclose enough information to recognize that the entity is a place. Similarly, the nouns \textit{pollution}, \textit{earthquake}, and \textit{temperatures} are commonly used to describe the state or events in a specific location. These findings are consistent with our initial intuition that specific words, along with their syntactic roles, can disclose important clues to recognize a given entity.
\subsection{Error analysis}

\begin{table}[t!]
\centering
\small
\renewcommand{\arraystretch}{1.2}
\setlength{\tabcolsep}{7.4pt}
\begin{tabular}{llll} 
\toprule
\textbf{Dataset} & \textbf{Dep. Trees} & \textbf{F1} & \textbf{Affected F1} \\ \midrule 
SemEval-10       & Manual       & 86.49            & 83.18 (3.31$\downarrow$) \\
CoNLL-12         & Converted    & 88.97            & 86.94 (2.03$\downarrow$) \\
ON 5.0 (BN)      & Converted    & 89.22            & 86.92 (2.30$\downarrow$) \\
\bottomrule
\end{tabular}
\caption{Impact on the performance when the dependency trees are generated from scratch.
F1 shows the score using dependency trees from either manual annotation or converted from parsing trees. Affected F1 provides the scores with tool-generated dependency trees. 
}
\label{tab:tree_errors}
\end{table}
Our model captures reasonably well the intended syntactic patterns, but the performance greatly relies on the quality of the dependency trees. 
We assess the impact of the dependency trees by evaluating the model using automatically generated trees from the Stanford CoreNLP tool \cite{manning-EtAl:2014:P14-5}. We replace the trees in the validation set of the SemEval 2010 dataset, which originally has manually annotated dependency parses. Not surprisingly, the F1 score drops by 3.31 absolute points (see Table \ref{tab:tree_errors}). We perform the same evaluation on OntoNotes 5.0 (BN) and CoNLL 2012. 
Similarly, the F1 score decreases by 2.30 and 2.03 points on each dataset, respectively. By further inspection of the trees generated from scratch, we find that they contain multiple roots mainly because of the multiple utterances that are transcribed from speech in the elaboration of these datasets. Multi-rooted trees affect the performance of the Tree-LSTM model because they prevent connecting information across the sentence. 
Additionally, Table \ref{tab:tree_errors} shows a smaller drop in performance on CoNLL 2012 and OntoNotes 5.0 (BN) compared to the drop on SemEval 2010. This is because the automatic conversions of the parsing trees do not have the same quality of the manual annotations in SemEval 2010 (e.g., multi-rooted sentences or erroneous connections between the nodes), and consequently the drop on SemEval 2010 tends to be bigger. 

\section{Conclusion}

We propose a novel approach that combines sequential and recursive linguistic properties for NER. Our model uses syntactic and semantic features extracted from \textsc{Tlstm} and \textsc{Blstm}, respectively. Then, we feed this information into the relative and global attention mechanisms. The relative attention allows our model to exploit linguistic properties over the sentence with respect to every word, while the global attention combines those properties to balance semantic and syntactic patterns. We benchmark our model on four datasets and establish a new state of the art on two of them.
By exploring the relationships among the entities and the most attended words, we find that the model learns to detect words that disclose information of the entity types based on their syntactic properties.


\bibliography{acl2019}
\bibliographystyle{acl_natbib}


\clearpage
\appendix

\section{Hyper-parameter tuning} \label{app:sec:hyperparameters}

We initialize the embedding matrices for part-of-speech tags and dependency relations using Xavier initialization \cite{glorot2010understanding}. While these embedding matrices are learned from scratch, we use the pre-trained language model ELMo to represent words \cite{peters-EtAl:2018:N18-1}. The size of the POS tag and dependency relation matrices are one fourth of the dimensions of the word vectors. We use the small version of ELMo in most of the experiments, which contains 256-dimensional word vectors. In this case, POS tag and dependency relation representations would be of 64 dimensions each.

We train all our models for 150 epochs using an initial learning rate of 5e-3 and a batch size of 16. The optimization of the models is conducted using Stochastic Gradient Descent (SGD) with momentum \cite{pmlr-v28-sutskever13}. We modify the learning rate every epoch using cosine annealing and warm restarts with $t_o = 10$ and a multiplicative factor of 2 \cite{DBLP:journals/corr/LoshchilovH16a}. Besides adding $\ell_2$ regularization, we also prevent over-fitting by applying input variational dropout \cite{Gal2015DropoutB} to every component in our model.

We consistently use gradient clipping among our experiments. We clip the norm of the gradient at 5.0 \citep{DBLP:journals/corr/abs-1211-5063, Goodfellow-et-al-2016}
$$\mathbf{g} \leftarrow \frac{\mathbf{g}\tau}{||\mathbf{g}||} ~~\mathrm{if}~ ||\mathbf{g}|| > \tau$$

To regularize the models, we use input variational dropout \citep{Gal2015DropoutB} choosing drop probabilities between 0.1 and 0.5, being 0.3 the best. We apply an $\ell_2$ regularization with a $\lambda$ penalty of 1e-5.

\section{Datasets} \label{app:sec:datasets}

\subsection{CoNLL 2003 Dataset}

The CoNLL 2003 dataset contains 4 entity labels: \texttt{PERSON}, \texttt{LOCATION}, \texttt{ORGANIZATION} and \texttt{MISCELANEOUS}. Table \ref{app:tab:conll03_general_stats} shows statistics on the entity labels.

\begin{table}
\centering
\small
\begin{tabular}{llll}
\toprule
Statistics		& Train		& Dev		& Test \\\midrule
Posts			& 14,041	& 3,250		& 3,453 \\
Tokens			& 203,621	& 51,362	& 46,435 \\
NE tokens		& 3,403		& 8,603		& 8,112 \\
NE tokens (\%)	& 1.67		& 16.74		& 17.46 \\
Uniqueness (\%)	& 26		& 40		& 41 \\\midrule
Classes			& 			& 			&    \\\midrule
Person			& 11,128	& 3,149		& 2,773 \\
Location		& 8,297		& 2,094		& 1,925 \\
Organization	& 10,025	& 2,092		& 2,496 \\
Miscellaneous	& 4,593		& 1,268		& 918 \\
\bottomrule
\end{tabular}
\caption{General statistics and class distribution of the CoNLL 2003 dataset. }
\label{app:tab:conll03_general_stats}
\end{table}

\subsection{SemEval 2010 Dataset}

The SemEval 2010 Task 1 dataset is a subset of the OntoNotes 5.0 corpus that contains manually annotated dependency parses. This is the main dataset for our approach because of the manual annotations. The dataset contains 3,648 sentences for training, 741	for development and 1,141 for testing. The entity labels on this dataset are shown in Table \ref{app:tab:semeval10_general_stats}.

\begin{table}
\centering
\small
\begin{tabular}{llll}
\toprule
\bf Entities 	& \bf Train & \bf Dev 	& \bf Test \\ \midrule
ANIMAL		&    9	&    0	&	  1  \\
CARDINAL	&  532	&   93	&	207  \\
DATE		& 1208	&  190	&	362  \\
DISEASE		&   28	&    1	&	  0 \\
EVENT		&   26	&    6	&	  6  \\
FAC			&   72	&   19	&	 16  \\
GAME		&    4	&    1	&	  0 \\
GPE			& 1244	&  259	&	405  \\
LANGUAGE	&    3	&    1	&	  8  \\
LAW			&   16	&    6	&	  8  \\
LOC			&  117	&   20	&	 39  \\
MONEY		&  296	&   57	&	 54  \\
NORP		&  588	&  110	&	205  \\
ORDINAL		&  120	&   28	&	 27  \\
ORG			& 1351	&  381	&	560  \\
PERCENT		&  332	&   27	&	 48  \\
PERSON		& 1282	&  331	&	372  \\
PLANT		&    8	&    1	&	  0 \\
PRODUCT		&   74	&   14	&	 57  \\
QUANTITY	&   52	&    4	&	 18  \\
SUBSTANCE	&  154	&   15	&	 31  \\
TIME		&  109	&   33	&	 50  \\
\bottomrule
\end{tabular}
\caption{Statistics and distribution of the classes in the SemEval 2010 dataset. }
\label{app:tab:semeval10_general_stats}
\end{table}

\subsection{CoNLL 2012 Dataset}

The CoNLL 2012 uses the entire OntoNotes 5.0 corpus and proposes a standard split of the data. Table \ref{app:tab:conll12_general_stats} shows the distribution of the labels after the split. Since the data comes from OntoNotes 5.0, the data contain parsing trees. Table \ref{app:tab:conll12_general_stats} shows the distribution of the labels.

\begin{table}
\centering
\small
\begin{tabular}{llll}
\toprule
\bf Entities & \bf Train & \bf Dev 	& \bf Test \\ \midrule
CARDINAL     &  10901  &    1720  & 1005  \\
DATE         &  18791  &    3208  & 1787  \\
EVENT        &   1009  &     179  & 85  \\
FAC          &   1158  &     133  & 149  \\
GPE          &  21938  &    3649  & 2546  \\
LANGUAGE     &    355  &      35  & 22  \\
LAW          &    459  &      65  & 44  \\
LOC          &   2160  &     316  & 215  \\
MONEY        &   5217  &     853  & 355  \\
NORP         &   9341  &    1277  & 990  \\
ORDINAL      &   2195  &     335  & 207  \\
ORG          &  24163  &    3798  & 2002  \\
PERCENT      &   3802  &     656  & 408  \\
PERSON       &  22035  &    3163  & 2134  \\
PRODUCT      &    992  &     214  & 90  \\
QUANTITY     &   1240  &     190  & 153  \\
TIME         &   1703  &     361  & 225  \\
WORK OF ART  &   1279  &     202  & 169  \\
\bottomrule
\end{tabular}
\caption{Statistics and distribution of the classes in the CoNLL 2012 dataset. }
\label{app:tab:conll12_general_stats}
\end{table}

\subsection{OntoNotes 5.0 Broadcast News Dataset}

The broadcast news section of the OntoNotes 5.0 corpus is a small portion of the data. Table \ref{app:tab:ontonotes_general_stats} shows the distribution of the labels.

\begin{table}
\centering
\small
\begin{tabular}{llll}
\toprule
\bf Entities 	& \bf Train & \bf Dev 	& \bf Test \\ \midrule
CARDINAL       &  1639  &   155 &   196  \\
DATE           &  2351  &   302 &   318  \\
EVENT          &   111  &    14 &    24  \\
FAC            &   275  &    24 &    36  \\
GPE            &  4056  &   516 &   537  \\
LANGUAGE       &    22  &     7 &     5  \\
LAW            &    25  &     6 &     4  \\
LOC            &   373  &    42 &    60  \\
MONEY          &   177  &    14 &    20  \\
NORP           &  2394  &   244 &   304  \\
ORDINAL        &   366  &    53 &    47  \\
ORG            &  2468  &   303 &   264  \\
PERCENT        &   132  &    21 &     6  \\
PERSON         &  4242  &   557 &   460  \\
PRODUCT        &   327  &    35 &    43  \\
QUANTITY       &   126  &    19 &    16  \\
TIME           &   517  &    68 &    54  \\
WORK OF ART    &   160  &    26 &    35  \\
\bottomrule
\end{tabular}
\caption{General statistics and class distribution of the OntoNotes 5.0 Broadcast News section dataset. }
\label{app:tab:ontonotes_general_stats}
\end{table}









\end{document}